\documentclass{article}

     \usepackage[final]{neurips_2020}

\usepackage[utf8]{inputenc} 
\usepackage[T1]{fontenc}    
\usepackage{hyperref}       
\usepackage{url}            
\usepackage{booktabs}       
\usepackage{amsfonts}       
\usepackage{nicefrac}       
\usepackage{microtype}      
\usepackage{bm}
\title{Pareto-efficient Acquisition Functions \\for Cost-Aware Bayesian Optimization}

\author{%
Gauthier Guinet\thanks{Joint first author.}~~\thanks{Work done during an internship at Amazon Web Services.} \\
  MIT\\
  \texttt{gguinet@mit.edu} \\
  \And
  Valerio Perrone$^*$ \\
  Amazon Web Services \\
  \texttt{vperrone@amazon.com} \\
  \And
  C\'edric Archambeau \\
  Amazon Web Services \\
  \texttt{cedrica@amazon.com} \\
}

\usepackage{microtype}
\usepackage{graphicx}
\usepackage{subfigure}
\usepackage{booktabs} 
\usepackage[numbers]{natbib}
\usepackage{xcolor}
\usepackage{amsmath}
\usepackage{amsfonts}
\usepackage{tikz}
\usepackage{url}
\usepackage{algorithm}
\usepackage{algorithmic}
\usepackage{enumitem}

\def\xb{{\mathbf x}}

\def\vec{{\mathrm{vec}}}

\newcommand{\cX}{\mathcal{X}}
\newcommand{\cD}{\mathcal{D}}
\renewcommand{\vec}[1]{\bm{#1}}

\usepackage[colorinlistoftodos,bordercolor=orange,backgroundcolor=orange!20,linecolor=orange,textsize=tiny]{todonotes}

\begin{document}

\maketitle

\begin{abstract}
Bayesian optimization (BO) is a popular method to optimize expensive black-box functions. It efficiently tunes machine learning algorithms under the implicit assumption that hyperparameter evaluations cost approximately the same. In reality, the cost of evaluating different hyperparameters, be it in terms of time, dollars or energy, can span several orders of magnitude of difference. While a number of heuristics have been proposed to make BO cost-aware, none of these have been proven to work robustly. In this work, we reformulate cost-aware BO in terms of Pareto efficiency and introduce the cost Pareto Front, a mathematical object allowing us to highlight the shortcomings of commonly used acquisition functions. Based on this, we propose a novel Pareto-efficient adaptation of the expected improvement. On 144 real-world black-box function optimization problems we show that our Pareto-efficient acquisition functions significantly outperform previous solutions, bringing up to 50\% speed-ups while providing finer control over the cost-accuracy trade-off. We also revisit the common choice of Gaussian process cost models, showing that simple, low-variance cost models predict training times effectively.
\end{abstract}

\section{Introduction}

Bayesian optimization (BO) is a well-established methodology to find the global minimizer of an expensive black-box function $f: \cX \rightarrow \mathbb{R}$, where $\cX \subset \mathbb{R}^d$~\citep{Shahriari2016}. One of the major use cases is the automatic tuning of machine learning algorithms~\cite{Hutter2011seqmod, Bergstra2011hpo, snoek2012practical, swersky2013multi, Shahriari2016}. BO has been successful in several other applications, including user interfaces~\cite{Brochu2010anima}, robotics~\cite{berkenkamp2016bayesian}, environmental monitoring~\cite{Marchant2012envir}, sensor networks~\cite{garnett2010bayesian}, adaptive Monte Carlo~\cite{wang2013adaptive}, experimental design~\cite{Azimi2012hyrbid}, and reinforcement learning~\cite{barsce2018autonomous, turchetta2019robust}. In all these settings, we typically do not have access to an analytic form of $f$ and only have a finite evaluation budget to sequentially query $f$ at points $\vec{x} \in \cX$.

A limitation of standard BO is the implicit assumption that evaluating different hyperparameter configurations incurs approximately the same cost, which is rarely the case in practice. For instance, evaluating hyperparameter settings corresponding to larger neural network architectures requires higher training cost. As cost can differ in orders of magnitudes~\cite{lee2020costaware}, some works have proposed heuristics to make BO \textit{cost-aware}~\cite{snoek2012practical, abdolshah2019costaware, lee2020costaware}. Unfortunately, cost-aware BO has not been studied systematically, and commonly used acquisition functions have not been proved to work robustly.

In this paper, we revisit cost-aware BO through the lens of Pareto optimality. This allows us to highlight the shortcomings of commonly-used heuristics and directly control the improvement-cost trade-off at each BO iteration. Based on this, we introduce a novel, robust cost-aware acquisition function. In extensive experiments drawn from 144 real-world hyperparameter optimization (HPO) problems, we show that our solution outperforms both classical and recent cost-aware methods. We also show that simple linear models predict training times more effectively than GPs, the default choice in the literature. 
\section{Background and Related Work}
Consider the problem of finding $\bm{x}_{\star} = \mathrm{argmin_{\bm{x}\in \mathcal{X}}}\ f(\bm{x})$, where $\bm{x}$ is a hyperparameter configuration,  $\mathcal{X}$ the search space, and  $f(\bm{x})$ a black-box function (e.g., the map from hyperparameter configurations to validation error). As the functional form of $f(\bm{x})$ is unknown, BO replaces it with a probabilistic surrogate model, with a popular choice being the Gaussian process (GP)~\citep{Rasmussen2006}. The next point to evaluate is obtained by repeatedly optimizing an \emph{acquisition function} until some budget (e.g., iterations, dollars or time) is exhausted~\citep{Jones1998:Efficient}. 

Most acquisition functions used in BO implicitly assume that all hyperparameter evaluations cost approximately the same. For example, the Expected Improvement (EI), one of the most popular acquisition function for BO, is defined by $EI(\vec{x}) = \mathbb{E}[\max(0, f_{min} - f(\vec{x} | \cD)]$ for each hyperparameter configuration $\vec{x}\in \cX$ \cite{Mockus1978:Application}. To make the EI cost-aware, common practice is to normalize it by the cost $c(\vec{x})$ of evaluating $\vec{x}$~\cite{snoek2012practical, poloczek2017multi, swersky2013multi}. This replaces EI with \textit{EI per unit cost (EIpu)} defined as follows: $$EIpu(\vec{x}) = \frac{EI(\vec{x})}{c(\vec{x} )}.$$ The modified form of EI is designed to balance the improvement with the evaluation cost~\cite{snoek2012practical}. To compute EIpu, it is necessary to learn the cost function $c(\xb)$, which is typically modeled with a warped GP \cite{snelson2004warped} fitted on the log cost $\gamma(\xb)$~\cite{snoek2012practical}. However, in \cite{lee2020costaware} it was shown that EIpu does not consistently improve over EI, especially when the best hyperparameters are in the most expensive regions of the hyperparameter space. To mitigate this, that work introduced a cost-cooling approach. If $\tau_k$ of the total budget $\tau$ has been used at the $k$th BO iteration (at $k=0$, $\tau_k = \tau_{init}$), EI-cool is defined as $EI_{\text{cool}}^{k}(\vec{x}) = \frac{EI(\vec{x})}{c(\vec{x})^\alpha} \;,\; \alpha = (\tau - \tau_k) / (\tau - \tau_{init})$. As the parameter $\alpha$ decays from one to zero, EI-cool transitions from EIpu to EI. As a result, cheap evaluations are obtained before expensive ones. While more robust than EIpu, experimental results in \cite{lee2020costaware} still only showed modest improvements over conventional EI. Moreover, this method requires the budget $\tau$ to be defined \emph{a priori}, with huge yet hard to study impact on performance. Other acquisition functions, such as entropy-based ones, are not designed to account for evaluation cost either~\cite{Hennig2012:Entropy,Hernandez2014:Predictive,Wang2017:Max}. 

Alternative approaches to cost-aware BO operate in a \textit{grey-box} setting~\cite{forrester2007multi, kandasamy2017multi, poloczek2017multi, wu2019practical}. Among these, multi-fidelity methods, such as Hyperband \cite{li2017hyperband} and its BO extensions \cite{falkner2018bohb, klein2017fastbo}, assume the presence of a fidelity parameter $s$. An example in the context of tuning neural networks is the epoch count. This parameter acts as a noisy proxy for high-fidelity evaluations in that increasing $s$ decreases noise at the expense of runtime. Except when $s$ is chosen to be the dataset subsampling factor, these methods are not applicable in a black-box setting as they require an algorithm-specific fidelity parameter $s$. In addition, by relying on parallel computing, multi-fidelity techniques target time-efficiency rather than compute time, cost, or energy efficiency. Existing sample-efficient, BO-based extensions of Hyperband are still built on cost-unaware acquisition functions such as the EI, and could be combined with the black-box, cost-aware approaches we propose in this work. Another grey-box, cost-aware technique was recently proposed in the context of multi-objective BO~\cite{abdolshah2019costaware}. 
Yet, this requires as an input a partial order of the search space dimensions based on \textit{a priori} cost preferences, a rather restrictive condition in practice.

\section{Pareto Efficient Bayesian Optimization}
\label{sec:pareto_eff_BO}
Without loss of generality, assume that cost is the time required to evaluate the black-box function $f$ at a given BO iteration. This is easily mapped to energetic or financial cost~\cite{hutter2012algorithm}. Assume BO is run for $N$ iterations and let $\mathcal{BO}_{\bm{\mathcal{A}}}(\cX^{N})$ denote the set of all hyperparameter configurations queried by BO using an acquisition function from class $\bm{\mathcal{A}}$. 

We aim to address the following problem: \emph{Given some budget, how do we develop a BO algorithm that uses it optimally?} Based on how the budget is defined, this maps to either one of the two following sub-problems:
\begin{itemize}
	\item \textbf{Bi-Optimization Problem.} The budget is the maximum number of BO iterations $N$. The goal is to find an optimal trade-off between the accuracy of the returned solution and the time required by BO. Formally, we want to determine the following Pareto Front:
	\begin{equation*}
	    \bm{\mathcal{PF}}(\{(y^{\star},c^{\star})\mid \vec{x}_{i}\in \mathcal{BO}_{\bm{\mathcal{A}}}(\cX^{N}),\; y^{\star} = \min_{1\leq i \leq N}f(\vec{x}_{i}),\; c^{\star}={\textstyle\sum_{i=1}^{N}}c(\vec{x}_{i}) \}) .
	\end{equation*}
	\item \textbf{Optimal Time Allocation Problem.} The budget is the maximum wall-clock time $\tau$. The goal is to maximize accuracy within the time budget $\tau$, with no constraints on the number of iterations. In other words, this setting is a limited-resource allocation or constrained optimization problem. Formally, we are interested in
	\begin{equation*}
	    \inf_{N\in \mathbb{N}^{*}}\{y^{\star}\mid \vec{x}_{i}\in \mathcal{BO}_{\bm{\mathcal{A}}}(\cX^{N}),\;y^{\star} = \min_{1\leq i \leq N}f(\vec{x}_{i}),\;\textstyle\sum_{i=1}^{N}c(\vec{x}_{i})\leq \tau\} .
	\end{equation*}
\end{itemize}
Next, we introduce a family $\bm{\mathcal{A}}_{\lambda}$ of Pareto-efficient acquisition functions to address these problems.

\subsection{Pareto Efficient Expected Improvement}

\begin{figure}[t]
	\centering
	\includegraphics[width = 0.49\textwidth]{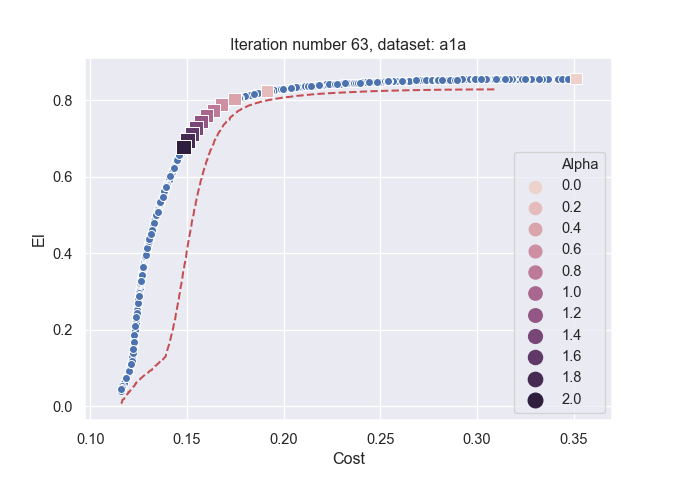}
	\includegraphics[width = 0.49\textwidth]{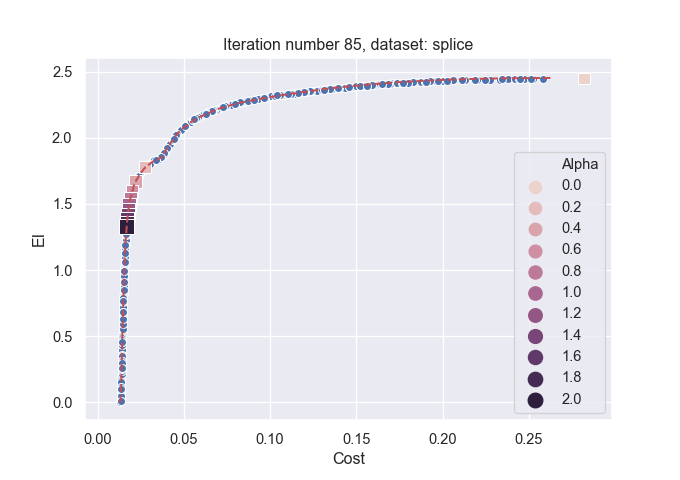}
\vspace{-0.4cm}
	\caption{Representative examples of the EI-cost Pareto front at two different BO iterations. XGBoost is tuned on the \texttt{a1a} and \texttt{splice} datasets from the UCI machine learning repository~\cite{Dua:2019}. The squares correspond to the maximizer of $EI_\alpha$ for different values of $\alpha$; recall that $\alpha = 0$ corresponds to EI and $\alpha = 1$ to EIpu. The blue dots represent the Pareto front at the current BO iteration $t$, while the red dashed curve refers to the Pareto front at iteration $t-1$.}.
	\label{fig:paretofront}
\end{figure}

Let $g: \cX \rightarrow \mathbb{R}^2$ be a function over $\cX$ mapping hyperparameters to the evaluation cost and negative EI. Given two points $\vec{x}_1, \vec{x}_2 \in \cX$, $\vec{x}_1 \succeq \vec{x}_2$ if $\vec{x}_2$ is \emph{weakly dominated} by $\vec{x}_1$, namely if and only if $g(\vec{x}_1)_i \leq g(\vec{x}_2)_i$ for $i \in \{1,2\}$. We write $\vec{x}_1 \succ \vec{x}_2$ if $\vec{x}_2$ is \emph{dominated} by $\vec{x}_1$, namely if and only if $\vec{x}_1 \succeq \vec{x}_2$ and $\exists i \in \{1,2\}$ such that $g(\vec{x}_1)_i < g(\vec{x}_2)_i$. The Pareto front of $g$ is defined by $\bm{\mathcal{PF}} = \{\vec{x} \in \cX | \not \exists \vec{x}' \in  \cX: \vec{x}'  \succ \vec{x}\}$, that is, the set of all non-dominated points in terms of EI and cost. 
We argue that the next point to be evaluated at a given iteration should be in the Pareto front; otherwise, it would be possible to find a better point either in terms of cost or EI. 
To this end, we define the family $\bm{\mathcal{A}}_{\alpha}$ of acquisition functions \textit{$EI_{\alpha}$}, where
\begin{align}
EI_{\alpha}(\vec{x}) = \frac{EI(\vec{x})}{c(\vec{x})^\alpha} \;,\; \alpha \in \mathbb{R}^{+}.
\end{align}
In the context of multi-objective optimization, $EI_{\alpha}$ is a parametric scalarization technique \cite{multi_obj}. The parameter $\alpha$ controls the trade-off between cost and EI and allows us to navigate the Pareto front, as illustrated by Figure~\ref{fig:paretofront}. On the one hand, EI (i.e., $\alpha = 0$) picks the point with highest EI, which is necessarily in the Pareto front. However, the best point usually involves spending an extra supplementary budget for very little gain of EI due to a performance plateau. The marginal improvement from optimizing the EI is usually not worth the additional cost (see Appendix \ref{AppendixB} for more details). On the other hand, EIpu (i.e., $\alpha = 1$) maximizes the expected improvement per unit of cost. By construction, the selected point will again belong to the Pareto front: it will be the point in the Pareto front with the highest slope. Yet, this may lead to a sub-optimal solution in terms of the accuracy that could have been achieved. 

In order to achieve a better and more robust trade-off, we propose to leverage the Pareto front at each iteration to select a good $\alpha$. To this end, we introduce the family $\bm{\mathcal{A}}_{\lambda}$ of \emph{contextual EI} ($\operatorname{CEI}$) acquisition functions, a cost-aware adaptation of the EI which takes into account the Pareto front at each iteration. For $\lambda \in [0,1]$, CEI is defined as follows:

\begin{align}
\operatorname{CEI}_{\lambda}(\vec{x}) := \left\{
\begin{array}{ll}
-c(\vec{x})  & \mbox{if } EI(\vec{x}) \geq (1-\lambda)\max_{\vec{z} \in \cX}(EI(\vec{z})),  \\
- \infty & \mbox{otherwise.}
\end{array}
\right.
\end{align}

Intuitively, CEI minimizes the cost $c$ among the points with a sufficiently high expected improvement. CEI only considers the $100 \times \lambda$\% points with the highest accuracy and, among these, selects the one with the lowest cost i.e. optimizes the cost constrained by EI. It can be shown with continuity arguments that the solution obtained belongs to the Pareto Front. Similar to the $\alpha$ parameter in $EI_\alpha$, $\lambda$ controls the trade-off between cost and improvement. However, as we will show shortly, CEI is more robust if the shape of the Pareto Front is unconventional. 
It also reveals the core importance of cost models, as detailed in the next section. 

\subsection{Cost Modeling}
\label{sec:cost_modeling}
The problem of predicting the cost of a computer program is well-studied by prior work. Applications include predicting system loads, dispatching computational resources, or determining computational feasibility~\cite{huan2010exec, hutter2012algorithm, Di2014cost, priya2011pred, lee2020costaware}. In the black-box setting, the typical procedure is to model the cost $c(\xb)$ with a warped GP~\cite{snelson2004warped} that fits the log cost $\gamma(\xb)$~\cite{snoek2012practical}. However, GPs extrapolate poorly, leading to high-variance cost predictions far away from data \cite{lee2020costaware}. 

\begin{figure}[t]
	\centering
	\includegraphics[width = 0.49\textwidth]{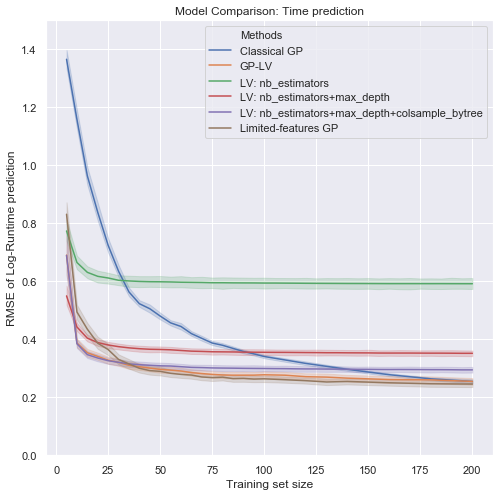}
	\includegraphics[width = 0.49\textwidth]{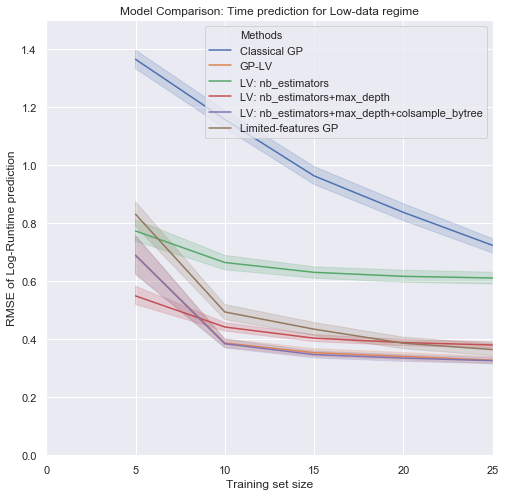}
	\caption{\textit{Left:} Comparison of models to predict the XGBoost training time as a function of its hyperparameters. The y-axis is the RMSE averaged over 144 HPO problems and 10 seeds, while the x-axis is the number of available hyperparameter evaluations to fit the cost model. We consider three simple linear models, using 1, 2 and 3 of the most significant hyperparameters, respectively. Limited-features GP is a GP trained using only 3 features. For all models, the features are log-scaled, and we predict the log run-time. \textit{Right:} Same experiment with a focus on the low-evaluation regime.
	}
	\label{fig:gp_lv}
\end{figure}

Consider the specific problem of modeling the evaluation time of XGBoost~\cite{Chen_2016} as a function of its hyperparameters. We compare different algorithms: a classical GP, linear (low-variance) models with different number of features (LV) \cite{lee2020costaware}, and a GP trained on the residuals of a low-variance model (GP-LV). These regression models need to be data-efficient to reach good accuracy quickly in the initial BO iterations. Figure~\ref{fig:gp_lv} shows that cost can be more efficiently captured by simple linear models, which we will use in the remainder of the paper. In the context of BO, the cost model is typically learned online as the optimization progresses. An alternative is to learn it using transfer learning from related tasks, as we detail in Appendix \ref{AppendixB}. 

\section{Experiments}
We considered 144 real-world black-box optimization problems, consisting of tuning XGBoost on 18 datasets, each processed with 8 different feature-engineering pipelines. Unless otherwise indicated, each result is averaged across the 144 problems and 10 seeds, with 95\% confidence intervals obtained via bootstrapping. The problems span multi-class, binary classification and regression tasks (more details in Appendix~\ref{AppendixA}). We ran all experiments on \texttt{AWS} with \texttt{m4.xlarge} machines.

\begin{wrapfigure}{r}{0.5\textwidth}
  \begin{center}
    \includegraphics[width=0.49\textwidth]{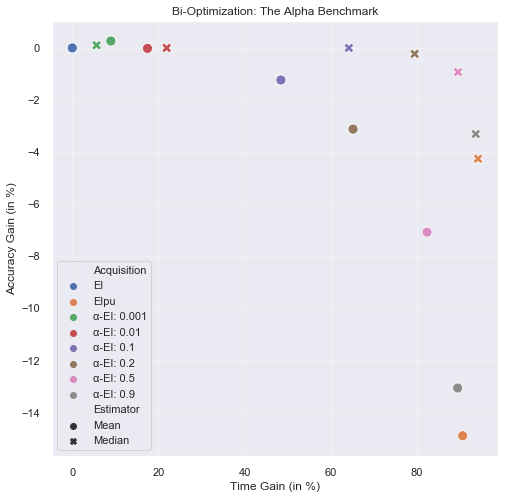}
  \end{center}
  \caption{The accuracy-cost trade off for $EI_\alpha$ at a set of $\alpha$ levels. Percentages are calculated w.r.t the performance of EI on the same blackbox problem and then averaged across seeds and datasets. Circles and crosses correspond respectively to the mean and median value all the blackboxes.}
	\label{fig:biopt_constant}
\end{wrapfigure}

\paragraph{Bi-optimization and Optimal Time Allocation}
We first study the problem of optimizing cost and performance through $EI_\alpha$ with a budget of 100 iterations. Figure~\ref{fig:biopt_constant} shows the impact of varying $\alpha$, which controls the trade-off between cost and accuracy. The resulting front appears smooth and the relationship between performance gain and cost can be observed. Particularly in terms of median, the Pareto front is sharp with large time gains traded for limited accuracy loss. Hence, it is possible to choose a value for $\alpha$ based on one's cost aversion. In particular, for low $\alpha$ values, the mean performance loss is limited compared to the cost gains: 50\% of wall-clock time is gained at 1\% accuracy loss (for $\alpha = 0.1$), and 20\% of wall-clock time is gained with no accuracy loss (for $\alpha = 0.01$). In contrast, standard practice of setting $\alpha=1$ (EIpu) leads to severe accuracy degradation, while $\alpha=0$ (EI) is unnecessarily costly.

\begin{figure}[b]
	\centering
	\includegraphics[width = 1\textwidth]{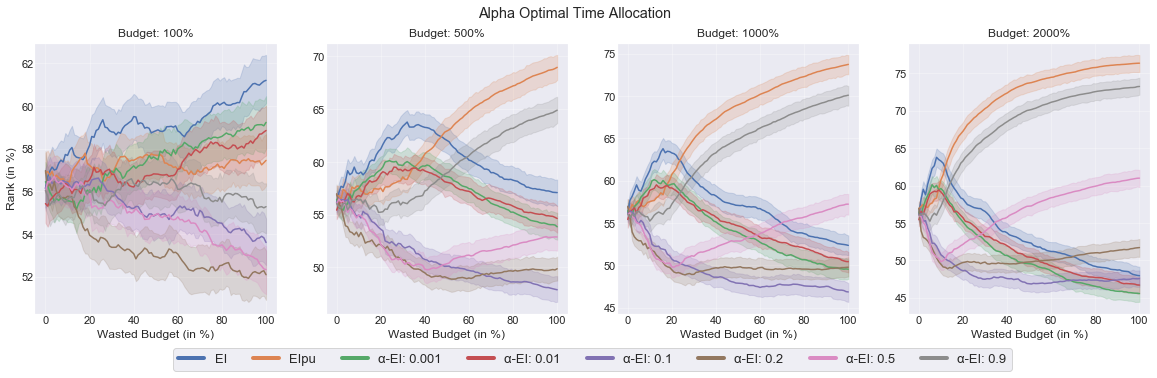}
	\caption{Comparison of $EI_\alpha$ with EI and EIpu in the optimal time allocation problem. Each plot corresponds to a different multiple of minimal budget (e.g. 2000\% is 20 times this budget). This notion is necessary to aggregate meaningfully results across datasets. Results are ranked at each iteration based on the minimum found up to that point by each method, a lower rank corresponding to a better minimization performance. This rank is then averaged across seeds and datasets.
	}
	\label{fig:optalloc_const}
\end{figure}

Next, we consider the problem of optimizing performance under a time constraint. As time scales are distinct across datasets, we consider the minimum time required by one of the $\alpha$-acquisitions function to reach 100 iterations. Figure~\ref{fig:optalloc_const} shows the results. Perhaps surprisingly, EIpu performs poorly even in a low-budget scenario. Generally, we observe a bell-shaped trend: $\alpha$ values close to 0 and 1 will produce less interesting results than intermediate values, with an optimum around 0.2. As we progressively increase the budget, EI progressively catches up with the different $EI_{\alpha}$; for larger budgets we are less sensitive to the actual cost and an excessive cost penalty degrades the performance as expected.

\begin{figure}[t]
	\centering
	\includegraphics[width = 0.49\textwidth]{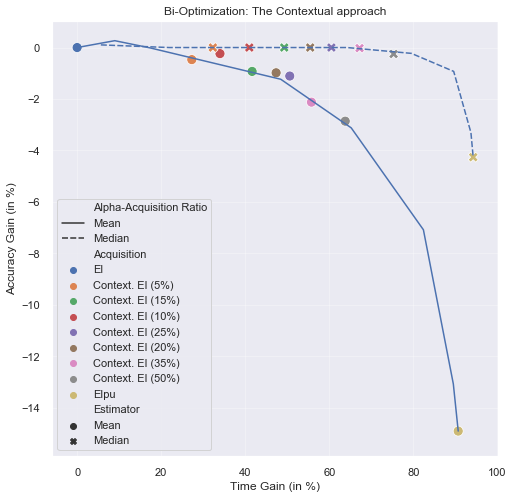}
	\includegraphics[width = 0.49\textwidth]{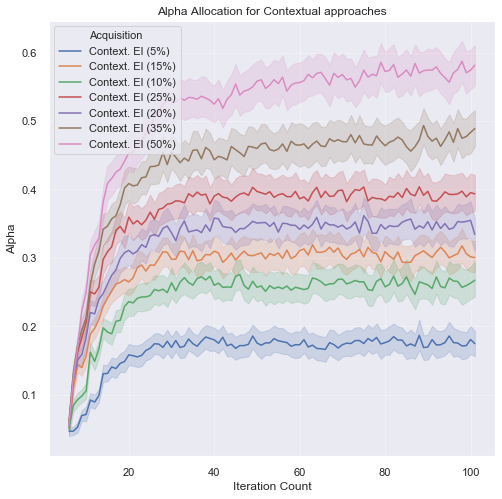}
	\caption{\textit{Left:} Bi-Optimization results for different contextual strategies. The percentage refers to the value of the trade-off ratio. The two blue curves correspond to fixed $\alpha$ strategies (mean and median). \textit{Right:} Average value of $\alpha$ picked by CEI against the number of BO iterations. 
	}
	\label{fig:biopt_context}
\end{figure}

\begin{figure}[t]
	\centering
	\includegraphics[width = 1\textwidth]{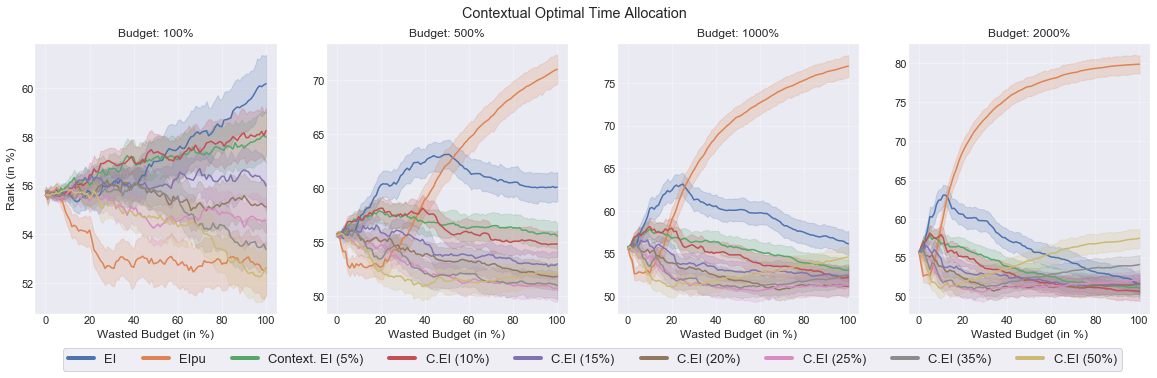}
	\caption{Comparison of CEI to the EI and EIpu acquisition functions in the optimal time allocation problem. 
	 The CEI acquisition functions consistently outperform the widely used EI and EIpu.
	}
	\label{fig:optalloc_context}
\end{figure}

The previous results and theoretical observations suggest allocating $\alpha$ dynamically rather than fixing it to a fixed value. We now demonstrate the benefits of adapting to the improvement-cost Pareto front via $\operatorname{CEI}$. Figure~\ref{fig:biopt_context} (left) shows that $\operatorname{CEI}$ is an effective solution to the bi-optimization problem, with $\lambda$ directly controlling the accuracy-cost trade-off. The trade-offs achieved by $\operatorname{CEI}$ are consistently on par or better compared to $EI_{\alpha}$. Indeed, for a fixed $\lambda$, $\operatorname{CEI}$ adapts to the Pareto front at each iteration, unlike $EI_{\alpha}$. Figure~\ref{fig:biopt_context} (right) reports the average $\alpha$ value that would have led to the configuration selected by $\operatorname{CEI}$, showing that $\operatorname{CEI}$ corresponds to dynamically allocating $\alpha$ at each iteration. Finally, we evaluate $\operatorname{CEI}$ in the context of the optimal time allocation problem, where the goal is to optimize performance within a fixed time budget. Figure~\ref{fig:optalloc_context} shows that, with the exception of extreme cost penalization, $\operatorname{CEI}$ consistently outperforms the popular EI and EIpu acquisitions. In particular, it is much more robust than EIpu and is able to outperform EI, even in the large budget regime.

\section{Conclusion}
\label{sec:ccl}
We introduced a novel formulation of cost-aware BO based on the Pareto front of cost and expected improvement. This allowed us to highlight the shortcomings of commonly used cost penalization heuristics and develop a new family of cost-aware acquisition functions based on EI. In extensive experiments we showed that our approach outperforms both the popular EI and its cost-aware extension. Future work could leverage the Pareto front dynamics during BO to develop more adaptive schedules for the cost-sensitivity parameter $\alpha$. Our method could also be applied to different acquisition functions and optimization problems.

\bibliographystyle{plain}
\bibliography{main.bib}

\newpage
\appendix
\section{Experiment settings}
\label{AppendixA}
Our code is built on GPyOpt~\cite{gpyopt2016}. Kernel hyperparameters for the GP models are obtained via maximum marginal likelihood estimation~\cite{Rasmussen2006}. We considered the problem of tuning the popular XGBoost algorithm (XGB)~\cite{Chen_2016}, as implemented in \texttt{scikit-learn}. This consists of a 7-dimensional hyperparameter space:  number of boosting rounds in $\{1, 2, \dots, 256\}$ (log scaled), learning rate in $[0.01, 1.0]$ (log scaled), minimum loss reduction to partition leaf node \texttt{gamma} in $[0.0, 0.1]$, L1 weight regularization \texttt{alpha} in $[10^{-3}, 10^3]$ (log scaled), L2 weight regularization \texttt{lambda} in $[10^{-3}, 10^3]$ (log scaled), subsampling rate in $[0.01, 1.0]$, maximum tree depth in $\{1, 2, \dots, 16\}$.

We ran all experiments on 144 benchmarks, consisting of the problem of tuning XGBoost on 18 datasets, each processed with 8 different feature-engineering pipelines. These pipelines involve basic operations, such as categorical variable detection, one-hot encoding, as well as dimensionality reduction operations, such as PCA.\footnote{The complete set of feature processing steps and their implementation is available at \url{https://github.com/aws/sagemaker-scikit-learn-extension}.} Results are averaged across 10 repetitions, with 95\% confidence intervals obtained via bootstrapping. The problems span multi-class, binary classification and regression tasks. We used the following publicly available datasets:
\begin{itemize}
    \item UCI datasets: \texttt{abalone}, \texttt{statloglandsatsatellite}, \texttt{turkiyestudentevaluation}, \texttt{insurancecompanybenchmarkcoil2000}, \texttt{parkinsonstelemonitoring}, \texttt{penbasedrecognitionofhandwrittendigits}.
    \item OpenML datasets: \texttt{3892, 405, 125922, 14953, 3007, 287, 503, 189, 558, 1489}.
    \item Kaggle datasets: \texttt{team-ai-spam-text-message-classification}, \texttt{olgabelitskaya-classification-of-handwritten-letters}.
\end{itemize}
\section{Additional Experiments}
\label{AppendixB}

The purpose of this appendix is two-fold. First, we give insights into the dynamics of cost and EI during BO. More precisely, we focus on its evolution and persistence. Second, we present additional results on cost learning and its impact on BO.

\paragraph{Evolution and Persistence of the Pareto Front}

We investigate the evolution of the Pareto front at each BO iteration. The difference in expected improvement across two subsequent BO iterations can be split as follows.

\begin{align*}
EI_{t+1}(\vec{x}) - EI_{t}(\vec{x}) &= \mathbb{E}_{t+1}[\max(0, y_{min}(t+1) - f(\vec{x} | \cD))] - \mathbb{E}_{t}[\max(0, y_{min}(t) - f(\vec{x} | \cD))] \\
												   &\simeq  \underbrace{y_{min}(t+1) -  y_{min}(t)}_{\textit{Constant \& Global}} + \underbrace{(\mathbb{E}_{t+1}-\mathbb{E}_{t})[ y_{min}(t) - f(\vec{x} | \cD)]}_{\textit{Non-linear \& Local}},
\end{align*} 
where $ \mathbb{E}_{t}$ stands for the expected value at iteration $t$, taken with respect to the posterior mean and variance functions at $t$, and $y_{min}(t)$ is the current objective function minimum. Yet, the magnitude and relative impact of these two components is experimentally unknown.

\begin{figure}[!h]
	\centering
	\includegraphics[width = 0.49\textwidth]{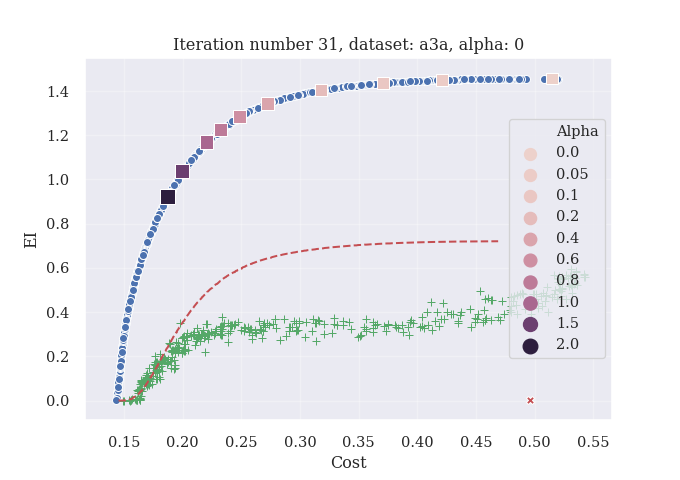}
	\includegraphics[width = 0.49\textwidth]{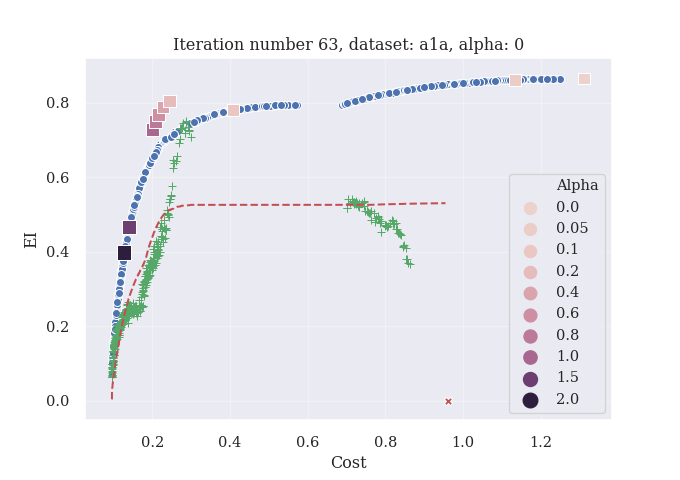}
	\includegraphics[width = 0.49\textwidth]{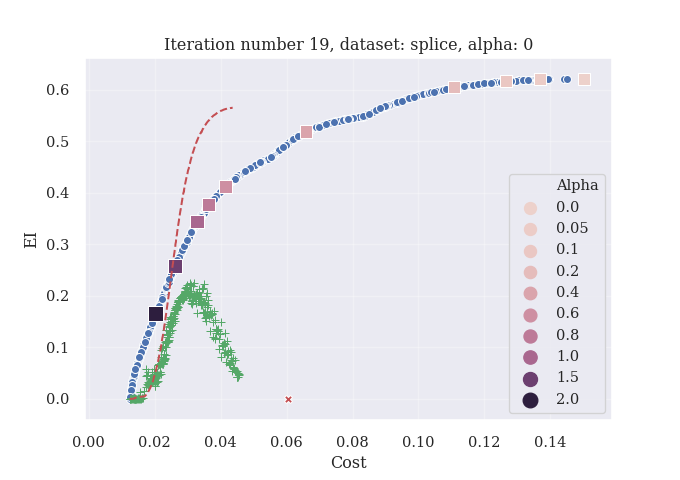}
	\includegraphics[width = 0.49\textwidth]{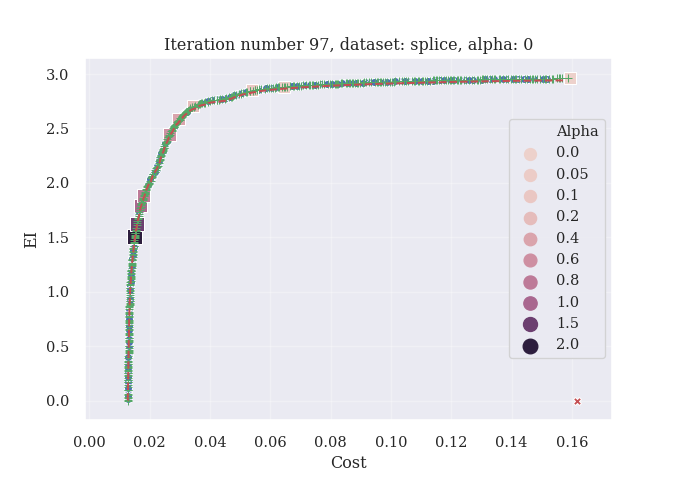}
	\caption{Representative examples of Pareto-Front visualization at different time-step $t$, focusing on its persistence across iterations. Each green cross corresponds to the new value of a point that belonged to the Pareto front at step $t-1$. The update is due to the fact that a new surrogate and cost model are used. Aside from this, legend and experiment settings are similar to Figure ~\ref{fig:paretofront}. The Pareto front remains stable only in the last example. XGBoost is tuned on the \texttt{a1a}, \texttt{a3a} and \texttt{splice} datasets from the UCI machine learning repository~\cite{Dua:2019} and we use the EI acquisition function. .}
	\label{fig:persistentparetofront}
\end{figure}

The Pareto front can drastically evolve across iterations. Figure~\ref{fig:persistentparetofront} shows this by comparing the Pareto front across two successive iterations, represented by the red and blue curves, respectively. This is mainly due to the non-linear and local component above. It is therefore complex to predict the state of the front at the next iteration. We observe the Pareto front tends to become stationary only for some HPO problems and after a large number of iterations. 

Then, we studied the persistence of the Pareto front. Specifically, we show that the optimal hyperparameter configurations rarely remain optimal across several evaluations. Figure~\ref{fig:persistentparetofront} illustrates this. This is expected due to the myopic nature of EI. Methods such as EIpu, which attempt to maximize a gain per cost, may suffer from this inconsistency. 

\paragraph{Cost Transfer Learning}

In Section \ref{sec:cost_modeling}, we presented results on how to build a better cost model during BO. To reduce the overhead of updating two models online, an alternative is learning the cost model offline in a transfer learning fashion. Although a similar idea was explored in \cite{priya2011pred}, learning a cost model across tasks has not been done in the context of BO. In terms of implementation, it is more practical to work with a fixed cost model, which does not evolve during BO. By learning a cost model offline, BO can immediately leverage an efficient cost-aware policy, without wasting any budget in an initial exploration phase. 

\begin{figure}[!h]
	\centering
	\includegraphics[width = 1\textwidth]{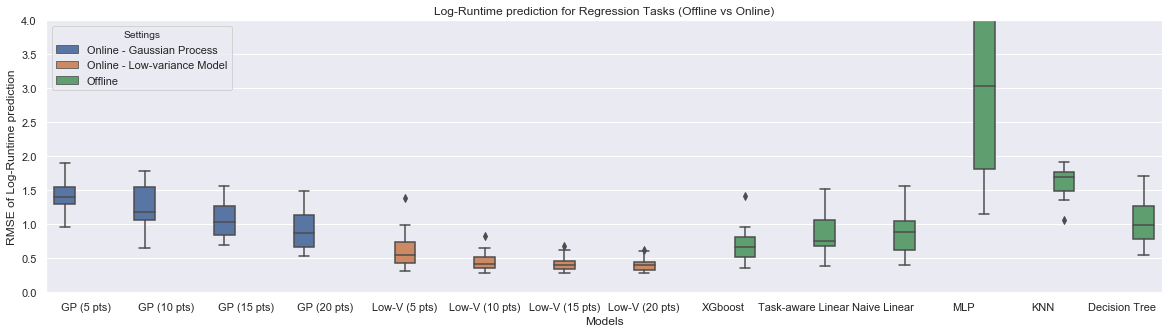}
	\includegraphics[width = 1\textwidth]{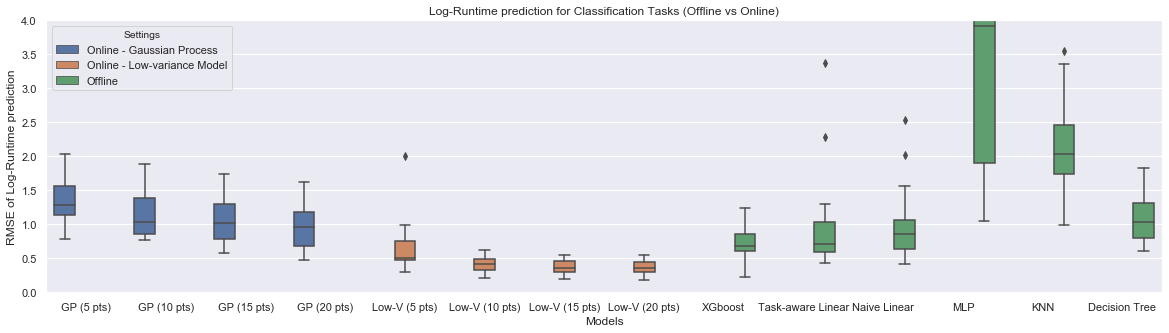}
	\caption{Comparison of different transfer learning and online cost models to predict the runtime of XGBoost as a function of its hyperparameters, averaged over 144 HPO problems. Results are split into regression and classification tasks. Blue boxes correspond to the classical GP and the orange ones to low-variance models using 3 features, with different training set sizes indicated in brackets. In green, different transfer learning regression algorithms are compared. We compare XGBoost, a low-variance linear model using data from related HPO problems (Task-Aware Linear), a LV model using all available data (Naive Linear), a multi-layer perceptron (MLP), a K-nearest neighbors regressor (KNN) and a Decision tree. Only core meta-features of datasets are used (number of classes, columns and lines) as supplementary features of the cost model. Yet, further experiments have shown that the impact of adding more complex meta-features is minor.}.
	\label{fig:offline}
\end{figure}

To demonstrate this, we compare offline and online models in Figure~\ref{fig:offline}. Although the cost model is easily captured by simple linear models, it does not generalize well across HPO problems. The models learned online achieve the same performance as the transfer learning models with only about ten hyperparameter evaluations. However, algorithms such as XGBoost or simple linear models can still model cost reasonably well despite not having access to any evaluations about the problem at hand. 

\paragraph{Impact of Cost Learning on BO Performance}

\begin{figure}[!h]
	\centering
	\includegraphics[width = 0.49\textwidth]{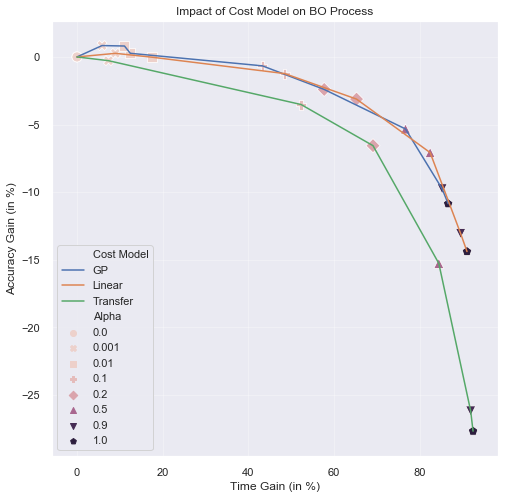}
	\caption{Comparison of different cost models when plugged into BO using $EI_\alpha$, for different values of $\alpha$ (Bi-optimization problem). Linear refers to the LV model with 3 features, and Transfer to the offline LV model, with the same 3 features.}
	\label{fig:cost_bo}
\end{figure}

We then evaluated the impact of the cost model on the overall BO performance. To this end, we focused on the bi-optimization problem and studied the cost-accuracy trade-off when plugging different cost models into BO (similar to \cite{lee2020costaware}). Figure~\ref{fig:cost_bo} shows that the cost model can have a strong impact on the performance-time gain. Transferring a cost model from offline datasets, which leads to a worse cost model, will also lead to a less interesting Pareto Front than the one obtained with a low-variance model learned online. More precisely, the time savings will be comparable but associated to higher accuracy losses. We also observe that the LV model performs at least as well as the GP model, at much lower computational complexity. In particular, for large values of $\alpha$, the LV model leads to a better Pareto Front.

\end{document}